\def\year{2019}\relax
\documentclass[letterpaper]{article} 
\usepackage{aaai18new}  
\usepackage{times}  
\usepackage{helvet}  
\usepackage{courier}  
\usepackage{url}  
\usepackage{graphicx}  
\usepackage{hyperref}
\usepackage{subcaption}
\newcommand{\md}[1]{\mathrm{#1}}             

\usepackage{amsmath} 
\usepackage{amssymb} 
\usepackage{natbib}

\begin{document}
%
\title{CrystalGAN: Learning to Discover Crystallographic Structures with Generative Adversarial Networks
}
\author{Asma Nouira$^1$, Nataliya Sokolovska$^2$, Jean-Claude Crivello$^1$\\
$^1$University Paris Est, ICMPE (UMR 7182) \\ CNRS, UPEC, F-94320 Thiais, France\\
$^2$Sorbonne University, INSERM, NutriOmics team, Paris France\\
}

\maketitle
\begin{abstract}

Our main motivation is to propose an efficient approach to generate novel multi-element stable chemical compounds that can be used in real world applications. 
This task can be formulated as a combinatorial problem, and it takes many hours of human experts to construct, and to evaluate new data.
Unsupervised learning methods such as Generative Adversarial Networks (GANs) can be efficiently used to produce new data. 
Cross-domain Generative Adversarial Networks were reported to achieve exciting results in image processing applications. 
However, in the domain of materials science, there is a need to synthesize data with higher order complexity compared to observed samples, and the state-of-the-art cross-domain GANs can not be adapted directly. 
In this contribution, we propose a novel GAN called CrystalGAN which generates new chemically stable crystallographic structures with increased domain complexity. 
We introduce an original architecture, we provide the corresponding loss functions, and we show that the CrystalGAN generates very reasonable data. 
We illustrate the efficiency of the proposed method on a real original problem of novel hydrides discovery that can be further used in development of hydrogen storage materials.

Keywords: Generative Adversarial Nets, Cross-Domain Learning, Materials Science, Higher-order Complexity.
\end{abstract}

\maketitle


\section{Introduction}

In modern society, a big variety of inorganic compositions are used for hydrogen storage owing to its favorable cost~\citep{Crivello16}. 
A vast number of organic molecules are applied in solar cells, such as organic light-emitting diodes, conductors, and sensors~\citep{Yang17}.
Synthesis of new organic and inorganic compounds is a challenge in physics, chemistry and in materials science. 
Design of new structures aims to find the best solution in a big chemical space, and it is in fact a combinatorial optimization problem.    

The number of applications of data mining methods in chemistry and materials science increases steadily~\citep{Seko18a}. 
There is a hope that recent developments in machine learning and data mining will accelerate the progress in materials science. 
Machine learning methods, namely, generative models, are reported to be efficient in new data generation~\citep{Friedman09}, and nowadays we have access to both, techniques to generate a huge amount of new chemical compounds, and to test the properties of all these candidates. 

In this work, we focus on applications of hydrogen storage, and in particular, we challenge the problem to investigate novel chemical compositions with stable crystals. 
Traditionally, density functional theory (DFT) plays a central role in prediction of chemically relevant compositions with stable crystals~\citep{Seko18}. 
However, the DFT calculations are computationally expensive, and it is not acceptable to apply it to test all possible randomly generated structures.

A number of machine learning approaches were proposed to facilitate the search for novel stable compositions~\citep{Butler18}. There was an attempt to find new compositions using an inorganic crystal structure database, and to estimate the probabilities of new candidates based on compositional similarities. 
These methods to generate relevant chemical compositions are based on recommender systems~\citep{Hu08}.
The output of the recommender systems applied in the crystallographic field is a rating or preference for a structure. 
A recent approach based on a combination of machine learning methods and the high-throughput DFT calculations allowed to explore ternary chemical compounds~\citep{Schmidt18}, and it was shown that statistical methods can be of a big help to identify stable structures, and that they do it much faster than standard methods.
Recently, support vector machines were tested to predict crystal structures~\citep{Oliynyk17} 
showing that the method can reliably predict the crystal structure given its composition. 
It is worth mentioning that data representation of observations to be passed to a learner, is critical, and data representations which are the most suitable for learning algorithms, are not necessarily scientifically intuitive~\citep{Swann18}.

Deep learning methods were reported to learn rich hierarchical models over all kind of data, and the GANs~\citep{Goodfellow14} 
is a state-of-the-art model to synthesize data. 
Moreover, deep networks were reported to learn transferable representations~\citep{Ren17}. 
The GANs were already exploited with success in cross-domain learning applications for image processing~\citep{Zhu17,Kim17,Janz17}.

\textit{Our goal is to develop a competitive approach to identify stable ternary chemical compounds, i.e., compounds containing three different elements, from observations of binary compounds.}

Nowadays, there does not exist any approach that can be applied directly to such an important task of materials science. 
The state-of-the-art GANs are limited in the sense that they do not generate samples in domains with increased complexity, e.g., the application where we aim to construct crystals with three elements from observations containing two chemical elements only. An attempt to learn many-to-many mappings was recently introduced by~\cite{Almahairi18}, however, this promising approach does not allow to generate data of a higher-order dimension.

Our contribution is multi-fold:
\begin{itemize}
\item To our knowledge, we are the first to introduce a GAN to solve the scientific problem of discovery of novel crystal structures, and we introduce an original methodology to generate new stable chemical compositions;
\item The proposed method is called \textit{CrystalGAN}, and it consists of two cross-domain GAN blocks with constraints integrating prior knowledge including a feature transfer step;  
\item The proposed model generates data with increased complexity with respect to observed samples;
\item We demonstrate by numerical experiments on a real challenge of chemistry and materials science that our approach is competitive compared to existing methods;
\item The proposed algorithm is efficiently implemented in Python, and it will be publicly available shortly.
\end{itemize}

This paper is organized as follows. 
First, we discuss the related work. 
Second, we provide the formalization of the problem, and introduce the CrystalGAN. 
The results of our numerical experiments are shown in the experimental section. Concluding remarks and perspectives close the paper. 

\section{Related Work}
\label{sec:relatedwork}

Our contribution is closely related to the problems of unsupervised learning and cross-domain learning, since our aim is to synthesize novel data, and the new samples are supposed to belong to an unobserved domain with an augmented complexity. 



In the adversarial nets framework, the deep generative models compete with an adversary which is a discriminative model learning to identify whether an observation comes from the model distribution or from the data distribution~\citep{Goodfellow16}. 
A classical GAN consists of two models, a generator $\mathbf{G}$ whose objective is to synthesize data and a discriminator $\mathbf{D}$ whose aim is to distinguish between real and generated data. 
The generator and the discriminator are trained simultaneously, and the training problem is formulated as a two-player minimax game.
A number of techniques to improve training of GANs were proposed by~\cite{Arjovsky17,Gulrajani17,Salimans16}. 

Learning cross domain relations is an active research direction in image processing. 
Several recent papers~\citep{Zhu17,Kim17,Almahairi18} 
discuss an idea to capture some particular characteristics of one image and to translate them into another image.
This problem is formalized as image-to-image translation, and there exist multiple applications, e.g., converting a grayscale image to a color image, or converting an image from one representation of a given scene to another.  
The state-of-the-art methods of~\cite{Zhu17,Kim17} 
are based on the property that the translation has to be cycle consistent. 
If a translator $\mathbf{G}: A \rightarrow B$ is used, then there exist another translator $\mathbf{F}: B \rightarrow A$ so that $\mathbf{G}$ and $\mathbf{F}$ are inverse of each other, and the mappings are bijective. 
The mappings $\mathbf{G}$ and $\mathbf{F}$ are trained simultaneously under the cycle consistency assumption what encourages $\mathbf{F}(\mathbf{G}(x)) \approx x$, and $\mathbf{G}(\mathbf{F}(x')) \approx  x'$. 
The objective function includes the adversarial losses on domains $A$ and $B$, and the cycle consistency loss. 

A conditional GAN for image-to-image translation is considered by~\cite{Isola17}. 
An advantage of the conditional model is that it allows to integrate underlying structure into the model. 
The conditional GANs were also used for multi-model tasks~\citep{Mirza14}.  
An idea to combine observed data to produce new data was proposed in~\citep{Yazdani17}, e.g., an artist can mix existing pieces of music to create a new one. 

An approach to learn high-level semantic features, and to train a model for more than a single task, was introduced by~\cite{Ren17}. 
In particular, it was proposed to train a model to jointly learn several complementary tasks. 
This method is expected to overcome the problem of overfitting to a single task. 
An idea to introduce multiple discriminators whose role varies from formidable adversary to forgiving teacher was discussed by~\cite{Durugkar17}. 

Several GANs were adapted to some materials science and chemical applications.  
So, Objective-Reinforced GANs that perform molecular generation of carbon-chain sequence taking into consideration some desired properties, were introduced in~\citep{Sanchez17}, and the method was shown to be efficient for drug discovery. 
Another avenue is to integrate rule-based knowledge, e.g., molecular descriptors with the deep learning. 
ChemNet~\citep{Goh17} is a deep neural network pre-trained with chemistry-relevant representations obtained from prior knowledge. 
The model can be used to predict new chemical properties. 
However, as we have already mentioned before, none of these methods generates crystal data of augmented complexity.

\begin{table*}
\centering
\footnotesize
\begin{tabular}{|ll|}
\hline
$A\md{H}$: & First domain, H is hydrogen, and $A$ is a metal \\
$B\md{H}$: & Second domain, H is hydrogen, and $B$ is another metal \\
$\mathbf{G}_{A\md{H}B_1}$: & Generator function that translates input features $x_{A\md{H}}$ from (domain) $A\md{H}$ to $B\md{H}$ \\
$\mathbf{G}_{B\md{H}A_1}$: & Generator function that translates input features $x_{B\md{H}}$ from (domain) $B\md{H}$ to $A\md{H}$ \\
$\mathbf{D}_{A\md{H}}$ and $\mathbf{D}_{B\md{H}}$: & Discriminator functions of  ${A\md{H}}$ domain and ${B\md{H}}$ domain, respectively \\
$A\md{H}B_1$: & $x_{A\md{H}B_1}$ is a sample generated by generator function $\mathbf{G}_{A\md{H}B_1}$ \\ 
$B\md{H}A_1$: & $y_{B\md{H}A_1}$ is a sample produced by generator function $\mathbf{G}_{B\md{H}A_1}$ \\
$A\md{H}BA_1$ and $B\md{H}AB_1$ : &  Data reconstructed after two generator translations \\
$A\md{H}B_g$ and $B\md{H}A_g$: & Data obtained after feature transfer step from domain $A\md{H}$ to domain $B\md{H}$, \\
& and from domain $B\md{H}$ to domain $A\md{H}$, respectively \\
& Input data for the second step of CrystalGAN \\
\hline
$\mathbf{G}_{A\md{H}B_2}$: & Generator function that translates  $x_{A\md{H}B_g}$ \\
& Features generated in the first step from $A\md{H}B_g$ to $A\md{H}B_2$ \\
$\mathbf{G}_{B\md{H}A_2}$: & Generator function that translates  $y_{B\md{H}A_g}$ \\
& Data generated in first step from $B\md{H}A_g$ to $B\md{H}A_2$ \\
$\mathbf{D}_{A\md{H}B}$ and $\mathbf{D}_{B\md{H}A}$: & The discriminator functions of domain ${A\md{H}B_g}$ and domain ${B\md{H}A_g}$, respectively\\
$A\md{H}B_2$: & $x_{A\md{H}B_2}$ is a sample generated by the generator function $\mathbf{G}_{A\md{H}B_2}$ \\
$B\md{H}A_2$: & $y_{B\md{H}A_2}$ is a sample produced by the generator function $\mathbf{G}_{B\md{H}A_2}$ \\
$A\md{H}BA_2$ and $B\md{H}AB_2$: & Data reconstructed as a result of two generators translations\\
$A\md{H}B_2$ and $B\md{H}A_2$: &  Final new data (to be explored by human experts) \\
\hline
\end{tabular}
\caption{Notations used in CrystalGAN.}
\label{tab:step1}
\end{table*}

\section{CrystalGAN: an Approach to Generate Stable Ternary Chemical Compounds}
\label{sec:method}

In this section, we introduce our approach. 
The CrystalGAN consists of three procedures:
\begin{enumerate}
\item \textit{First step GAN} which is closely related to the cross-domain GANs, and that generates pseudo-binary samples where the domains are mixed.
\item \textit{Feature transfer procedure} constructs higher order complexity data from the samples generated at the previous step, and where components from all domains are well-separated.
\item \textit{Second step GAN} synthesizes, under geometric constraints, novel ternary stable chemical structures.
\end{enumerate}
First, we describe a cross-domain GAN, and then, we provide all the details on the proposed CrystalGAN. 
We provide all notations used by the CrystalGAN in Table~\ref{tab:step1}. 
The GANs architectures for the first and the second steps are shown on Figure~\ref{fig:crystalgan}.

\subsection{A Cross-Domain GAN: Problem Formulation}

DiscoGAN~\citep{Kim17} and CycleGAN~\citep{Zhu17} propose a promising modification compared to the classic GAN: the model does not take the noise but samples from another domain, resulting in cross-domain learning.

We consider a function $\mathbf{G}_{ABZ}$ that maps elements from domains $A$ and $B$ to domain $Z$ which includes the co-domains $A$ and $B$. 
In an unsupervised learning scenario, $\mathbf{G}_{ABZ}$ can be arbitrarily defined, however, to apply it to real-world applications, some conditions on the relation of interest have to be well-defined. 

In an idealistic setting, the equality
\begin{align}
\mathbf{G}_{ABZ} \circ \mathbf{G}_{ZAB} (x_A, x_B) =  (x_A, x_B)
\end{align}
is satisfied. 
However, this constraint is a hard constraint, it is not straightforward to optimize it, and a relaxed soft constraint is preferred. 
As a soft constraint, we can consider the distance
\begin{align}
d\left(\mathbf{G}_{ABZ} \circ \mathbf{G}_{ZAB} (x_A, x_B), (x_A, x_B)\right), 
\end{align}
and minimize it using a metric function such as $L_1$ or $L_2$.

\begin{align}
- \mathbb{E}_{x_A, x_B \sim P_{A,B}} \left[ \log \mathbf{D}_{Z} (\mathbf{G}_{ABZ}) (x_A, x_B) \right].
\end{align}
The cross-domain GANs were shown to be efficient to discover relations between two different domains from unpaired samples, without any explicit labels, and to find a mapping from one domain to another. 
However, neither DiscoGAN, or CycleGAN are not able to generate data with increased complexity.

\subsection{Problem Formulation for Applications with Augmented Complexity}

We now propose a novel architecture based on the cross-domain GAN algorithms with constraint learning to discover higher order complexity crystallographic systems. 
We introduce a GAN model to find relations between different crystallographic domains, and to generate new materials.

To make the paper easier to follow, without loss of generality, we will present our method providing a specific example of generating ternary hydride compounds of the form "$A$ (a metal) - H (hydrogen) - $B$ (a metal)". 

The training algorithm observes stable binary compounds containing chemical elements $A$+H which is a composition of some metal~$A$ and the hydrogen~H, and $B$+H which is a mixture of another metal~$B$ with the hydrogen. 
So, a machine learning algorithm has access to observations $\{(x_{A\md{H}_i})\}_{i=1}^{N_{A\md{H}}}$ and $\{(y_{B\md{H}_i})\}_{i=1}^{N_{B\md{H}}}$. 
\textit{Our goal is to generate novel ternary, i.e. more complex, stable data $x_{A\md{H}B}$ (or $y_{B\md{H}A}$) based on the properties learned from the observed binary structures. }

\begin{figure}[hp!]
\centering
\begin{subfigure}[c]{.85\linewidth}
\includegraphics[scale=1.1]{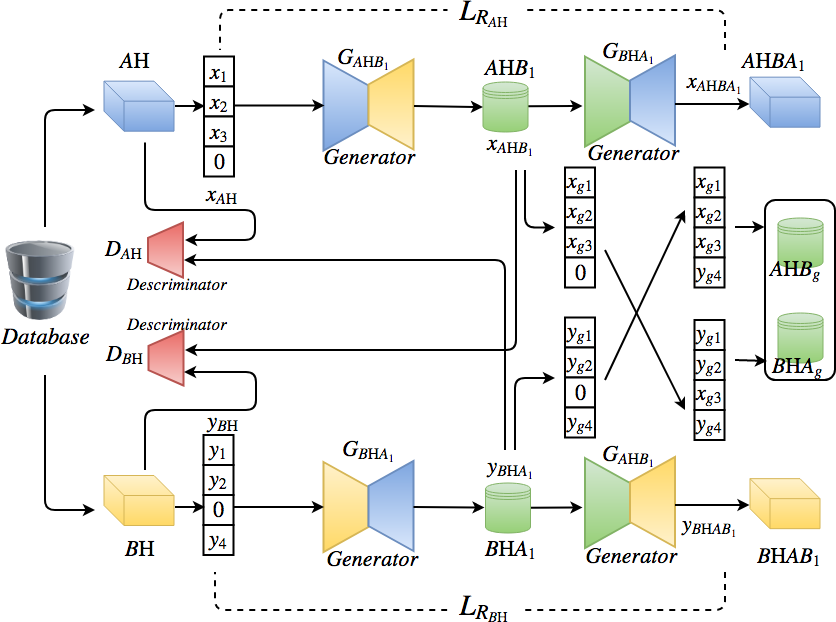}
\caption{First step of CrystalGAN.}
\label{fig:step1}
\vspace{1cm}
\end{subfigure}
\begin{subfigure}[c]{.75\linewidth}
\includegraphics[scale=1]{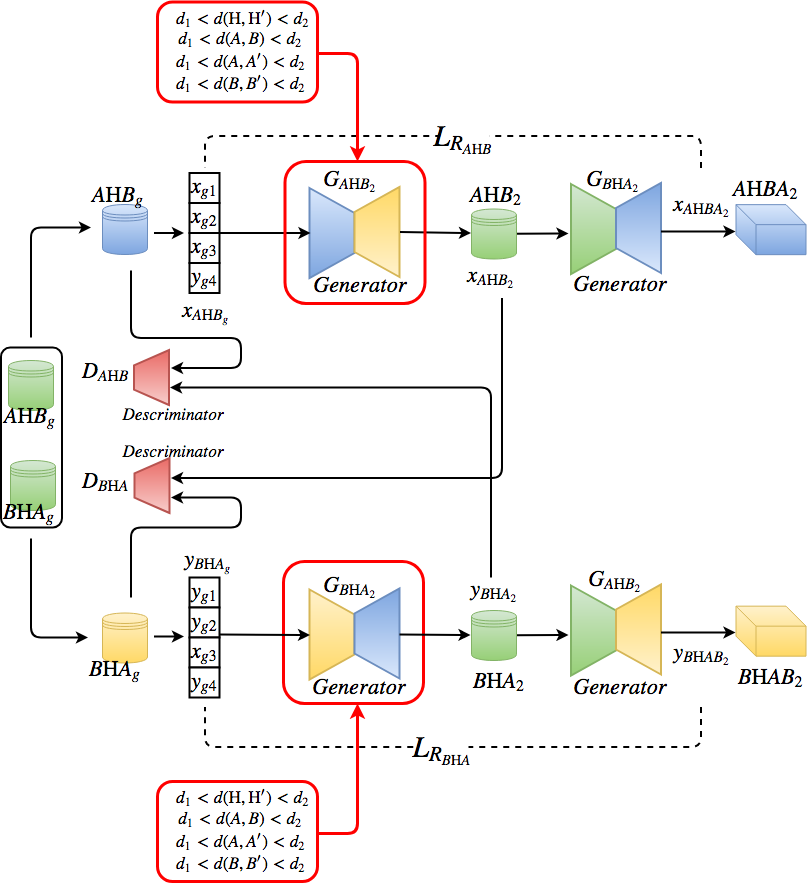}
\caption{Second step of CrystalGAN}
\label{fig:step2}
\end{subfigure}
\caption{The CrystalGAN architecture.}
\label{fig:crystalgan}
\end{figure}

We describe the architecture of the CrystalGAN on Figure~\ref{fig:crystalgan}.
\begin{figure}
\centering
\includegraphics[scale=0.3]{encodingdata.png}
\caption{Encoding of $x_{A\md{H}}$ and $y_{B\md{H}}$ with placeholders.}
\label{fig:samples}
\end{figure}
%
\subsection{Steps of CrystalGAN}

Our approach consists of two consecutive steps with a feature transfer procedure inbetween.

\subsubsection{First Step}

The first step of CrystalGAN generates new data with increased complexity. 
The adversarial network takes $\{(x_{A\md{H}_i})\}_{i=1}^{N_{A\md{H}}}$ and $\{(y_{B\md{H}_i})\}_{i=1}^{N_{B\md{H}}}$, and synthesizes 
\begin{align}
x_{{A\md{H}B}_1} = \mathbf{G}_{{A\md{H}B}_1}(x_{A\md{H}}),
\end{align}
\begin{align}
x_{{A\md{H}BA}_1} = \mathbf{G}_{{B\md{H}A}_1}(x_{{A\md{H}B}_1}) = \mathbf{G}_{{B\md{H}A}_1} \circ \mathbf{G}_{{A\md{H}B}_1}(x_{A\md{H}}).
\end{align}
and
\begin{align}
y_{{B\md{H}A}_1} = \mathbf{G}_{{B\md{H}A}_1}(y_{B\md{H}}),
\end{align}
\begin{align}
y_{{B\md{H}AB}_1} = \mathbf{G}_{{A\md{H}B}_1}(y_{{B\md{H}A}_1}) = \mathbf{G}_{{A\md{H}B}_1} \circ \mathbf{G}_{{B\md{H}A}_1}(y_{B\md{H}}).
\end{align}
Figure~\ref{fig:step1} summarizes the first step of CrystalGAN. 

The reconstruction loss functions take the following form:
\begin{align}
\mathcal{L}_{R_{A\md{H}}} = d(x_{{A\md{H}BA}_1},x_{A\md{H}}) =  d(\mathbf{G}_{B\md{H}A_1} \circ \mathbf{G}_{A\md{H}B_1}(x_{A\md{H}}), x_{A\md{H}}), \\
\mathcal{L}_{R_{B\md{H}}} = d(y_{B\md{H}AB_1},y_{B\md{H}}) =  d(\mathbf{G}_{A\md{H}B_1} \circ \mathbf{G}_{B\md{H}A_1}(y_{B\md{H}}), y_{B\md{H}}).
\end{align}

Ideally, $\mathcal{L}_{R_{A\md{H}}} = 0$, $ \mathcal{L}_{R_{B\md{H}}} = 0$,  and $ x_{A\md{H}BA_1} = x_{A\md{H}}$ , $ y_{B\md{H}AB_1} = y_{B\md{H}}$, and we minimize the distances $d(x_{A\md{H}BA_1},x_{A\md{H}}) $ and $ d(y_{B\md{H}AB_1},y_{B\md{H}})$.

The generative adversarial loss functions of the first step of CrystalGAN aim to control that the original observations are reconstructed as accurate as possible:
\begin{align}
\mathcal{L}_{\md{GAN}_{B\md{H}}} = - \mathbb{E}_{x_{A\md{H}} \sim P_{A\md{H}}}[\log (\mathbf{D}_{B\md{H}} (\mathbf{G}_{A\md{H}B_1} (x_{A\md{H}})))],
\end{align}
and
\begin{align}
\mathcal{L}_{\md{GAN}_{A\md{H}}} = - \mathbb{E}_{y_{B\md{H}} \sim P_{B\md{H}}} [\log (\mathbf{D}_{A\md{H}} (\mathbf{G}_{B\md{H}A_1} (y_{B\md{H}})))]. 
\end{align}

The generative loss functions contain the two terms defined above:

\begin{align}
\mathcal{L}_{\mathbf{G}_{A\md{H}B_1}} = \mathcal{L}_{\md{GAN}_{B\md{H}}} + \mathcal{L}_{R_{A\md{H}}},
\end{align}
\begin{align}
\mathcal{L}_{\mathbf{G}_{B\md{H}A_1}} = \mathcal{L}_{\md{GAN}_{A\md{H}}} + \mathcal{L}_{R_{B\md{H}}}.
\end{align}

The discriminative loss functions aim to discriminate the samples coming from $A\md{H}$ and $B\md{H}$:
\begin{align}
\mathcal{L}_{\mathbf{D}_{B\md{H}}} = &- \mathbb{E}_{y_{B\md{H}} \sim P_{B\md{H}}} [\log (\mathbf{D}_{B\md{H}}(y_{B\md{H}}))]  \\ \nonumber &-  \mathbb{E}_{x_{A\md{H}} \sim P_{A\md{H}}} [\log(1- \mathbf{D}_{B\md{H}}(\mathbf{G}_{A\md{H}B_1}(x_{A\md{H}})))],
\end{align}
\begin{align}
\mathcal{L}_{\mathbf{D}_{A\md{H}}} = &- \mathbb{E}_{x_{A\md{H}} \sim P_{A\md{H}}} [\log (\mathbf{D}_{A\md{H}}(x_{A\md{H}}))] \\ \nonumber &-  \mathbb{E}_{y_{B\md{H}} \sim P_{B\md{H}}} [\log(1- \mathbf{D}_{A\md{H}}(\mathbf{G}_{B\md{H}A_1}(y_{B\md{H}})))].
\end{align}
Now, we have all elements to define the full generative loss function of the first step:
\begin{align}
\mathcal{L}_{\mathbf{G_1}} &= \mathcal{L}_{\mathbf{G}_{A\md{H}B_1}} + \mathcal{L}_{\mathbf{G}_{B\md{H}A_1}} \\ \nonumber &=  \lambda_1 \mathcal{L}_{\md{GAN}_{B\md{H}}} + \lambda_2 \mathcal{L}_{R_{A\md{H}}} + \lambda_3 \mathcal{L}_{\md{GAN}_{A\md{H}}} + \lambda_4 \mathcal{L}_{R_{B\md{H}}},
\end{align}
where $\lambda_1$, $\lambda_2$, $\lambda_3$, and $\lambda_4$ are real-valued hyper-parameters that control the ratio between the corresponding terms, and the hyper-parameters are to be fixed by cross-validation.

The full discriminator loss function of this step $\mathcal{L}_{\mathbf{D_1}}$  is defined as follows:
\begin{align}
\mathcal{L}_{\mathbf{D_1}} = \mathcal{L}_{\mathbf{D}_{A\md{H}}} + \mathcal{L}_{\mathbf{D}_{B\md{H}}}.
\end{align}

\subsubsection{Feature Transfer}

The first step generates pseudo-binary samples $M\md{H}$, where $M$ is a new discovered domain merging $A$ and $B$ properties.
Although these results can be interesting for human experts, the samples generated by the first step are not easy to interpret, since the domains $A$ and $B$ are completely mixed in these samples, and there is no way to deduce characteristics of two separate elements coming from these domains.  

So, we need a second step which will generate data of a higher order complexity from two given domains.
We transfer the attributes of $A$ and $B$ elements, this procedure is also shown on Figure~\ref{fig:step1}, in order to construct a new dataset that will be used as a training set in the second step of the CrystalGAN.

In order to prepare the datasets to generate higher order complexity samples, we add a placeholder. 
(E.g., for domain $A\md{H}$, the fourth matrix is empty, and for domain $B\md{H}$, the third matrix is empty.) This implementation detail is sketched on Figure~\ref{fig:samples}.



\subsubsection{Second Step of the CrystalGAN}

The second step GAN takes as input the data generated by the first step GAN and modified by the feature transfer procedure. 
The results of the second step are samples which describe ternary chemical compounds that are supposed to be stable from chemical viewpoint. 
\textit{The geometric constraints control the quality of generated data.}

A crystallographic structure is fully described by a local distribution. 
This distribution is determined by distances to all nearest neighbors of each atom in a given crystallographic structure. 
We enforce the second step GAN with the following geometric constraints which satisfy the geometric conditions of our scientific domain application.
The implemented constraints are also shown on Figure~\ref{fig:step2}.

Let $S = \{s_i\}_{i=1}^m$ be the set of distances of the first neighbors of all atoms in a crystallographic structure. 
There are two geometric constraints to be considered while generating new data.

The first geometric (geo) constraint is defined as follows:
\begin{align}
 \mathcal{L}_{\md{geo}_1} = f(d_1, s_1, ..., s_m)
=  \min_{s \in S} 
\parallel d_1-s \parallel _2 ^2, 
\end{align}
where $d_1$ is the minimal distance between two first nearest neighbors in a given crystallographic structure.

The second geometric constraint takes the following form:
\begin{align} 
\mathcal{L}_{\md{geo}_2} = f(d_2, s_1, ..., s_m)  = - \min_{s \in S} \parallel d_2-s \parallel _2 ^2, 
\end{align}
where $d_2$ is the maximal distance between two first nearest neighbors.

The loss function of the second step GAN is augmented by the following geometric constraints:
\begin{align}
 \mathcal{L}_\md{geo} = \mathcal{L}_\md{geo_1} + \mathcal{L}_\md{geo_2} .
\end{align}

Given $x_{A\md{H}B_g}$ and $y_{B\md{H}A_g}$ from the previous step, we generate:
\begin{align}
x_{A\md{H}B_2} = \mathbf{G}_{A\md{H}B_2}(x_{A\md{H}B_g}),
\end{align}
\begin{align}
x_{A\md{H}BA_2} = \mathbf{G}_{B\md{H}A_2}(x_{A\md{H}B_2}) =  \mathbf{G}_{B\md{H}A_2} \circ \mathbf{G}_{A\md{H}B_2}(x_{A\md{H}B_g}).
\end{align}
and
\begin{align}
y_{B\md{H}A_2} = \mathbf{G}_{B\md{H}A_2}(y_{B\md{H}A_g}),
\end{align} 
\begin{align}
y_{B\md{H}AB_2} = \mathbf{G}_{A\md{H}B_2}(y_{B\md{H}A_2}) = 
\mathbf{G}_{A\md{H}B_2} \circ \mathbf{G}_{B\md{H}A_2}(y_{B\md{H}A_g}).
\end{align}
The reconstruction loss functions are given:
\begin{align}
\mathcal{L}_{R_{A\md{H}B}} &= d(x_{A\md{H}BA_2},x_{A\md{H}B_g}) \\ \nonumber &= d(\mathbf{G}_{B\md{H}A_2} \circ \mathbf{G}_{A\md{H}B_2}(x_{A\md{H}B_g}), x_{A\md{H}B_g}),
\end{align}
\begin{align}
 \mathcal{L}_{R_{B\md{H}A}} &= d(y_{B\md{H}AB_2},y_{B\md{H}A_g}) \\ \nonumber &=   d(\mathbf{G}_{A\md{H}B_2} \circ \mathbf{G}_{B\md{H}A_2}(y_{B\md{H}A_g}), y_{B\md{H}A_g}).
\end{align}
The generative adversarial loss functions are given by:
\begin{align}
\mathcal{L}_{\md{GAN}_{B\md{H}A_g}} =  - \mathbb{E}_{x_{A\md{H}B_g} \sim P_{A\md{H}B_g}} [\log (\mathbf{D}_{B\md{H}A} (\mathbf{G}_{A\md{H}B_2} (x_{A\md{H}B_g})))], \\
\mathcal{L}_{\md{GAN}_{A\md{H}B_g}} =  - \mathbb{E}_{y_{B\md{H}A_g} \sim P_{B\md{H}A_g}} [\log (\mathbf{D}_{A\md{H}B} (\mathbf{G}_{B\md{H}A_2} (y_{B\md{H}A_g})))].
\end{align}
The generative loss functions of the this step are defined as follows:
\begin{align}
 \mathcal{L}_{\mathbf{G}_{A\md{H}B_2}} = \mathcal{L}_{\md{GAN}_{B\md{H}A_g}} + \mathcal{L}_{R_{A\md{H}B}},
 \end{align}
\begin{align}
 \mathcal{L}_{\mathbf{G}_{B\md{H}A_2}} = \mathcal{L}_{\md{GAN}_{A\md{H}B_g}} + \mathcal{L}_{R_{B\md{H}A}}. 
\end{align}
The losses of the discriminator of the second step can be defined:
\begin{align}
\mathcal{L}_{\mathbf{D}_{B\md{H}A}} = & - \mathbb{E}_{y_{B\md{H}A_g} \sim P_{B\md{H}A_g}} [\log (\mathbf{D}_{B\md{H}A}(y_{B\md{H}A_g}))] \\ \nonumber - 
& \mathbb{E}_{x_{A\md{H}B_g} \sim P_{A\md{H}B_g}} [\log(1- \mathbf{D}_{B\md{H}A}(\mathbf{G}_{A\md{H}B_2}(x_{A\md{H}B_g})))], \\
\mathcal{L}_{\mathbf{D}_{A\md{H}B}} = &- \mathbb{E}_{x_{A\md{H}B_g} \sim P_{A\md{H}B_g}} [\log (\mathbf{D}_{A\md{H}B}(x_{A\md{H}B_g}))] \\ \nonumber - 
& \mathbb{E}_{y_{B\md{H}A_g} \sim P_{B\md{H}A_g}} [\log(1- \mathbf{D}_{A\md{H}B}(\mathbf{G}_{B\md{H}A_2}(y_{B\md{H}A_g})))].
\end{align}

Now, we have all elements to define the full generative loss function: 
\begin{align}
\mathcal{L}_{\mathbf{G_2}} = \mathcal{L}_{\mathbf{G}_{A\md{H}B_2}} + \mathcal{L}_{\mathbf{G}_{B\md{H}A_2}} + \mathcal{L}_\md{geo} \\ \nonumber =  
\lambda_1 \mathcal{L}_{\md{GAN}_{B\md{H}A_g}} + \lambda_2 \mathcal{L}_{R_{A\md{H}B}} +\lambda_3 \mathcal{L}_{\md{GAN}_{A\md{H}B_g}} + 
& \lambda_4 \mathcal{L}_{R_{B\md{H}A}} \\ \nonumber+ \lambda_5 \mathcal{L}_\md{geo_1} + \lambda_6 \mathcal{L}_\md{geo_2}, 
\end{align}
where $\lambda_1$, $\lambda_2$, $\lambda_3$, $\lambda_4$, $\lambda_5$, and $\lambda_6$ are real-valued hyper-parameters that control the influence of the terms.

The full discriminative loss function of the second step $\mathcal{L}_{\mathbf{D_2}} $ takes the form:
\begin{align}
 \mathcal{L}_{\mathbf{D_2}} = \mathcal{L}_{\mathbf{D}_{A\md{H}B}} + \mathcal{L}_{\mathbf{D}_{B\md{H}A}}.
\end{align}

To summarize, in the second step, we use the dataset issued from the feature transfer as an input containing two domains $x_{A\md{H}B_g}$ and $y_{B\md{H}A_g}$.
We train the cross-domain GAN taking into consideration constraints of the crystallographic environment.
We integrated geometric constraints proposed by crystallographic and materials science experts to satisfy environmental constraints, and to increase the rate of synthesized stable ternary compounds. 
The second step of CrystalGAN is drafted on Figure~\ref{fig:step2}.

\subsection{GAN Architecture}

A generator network is defined as $\mathbf{G}_{A\md{H}B_1}: \mathbb{R}_{A\md{H}}^{l \times m}, \mathbb{R}_{B\md{H}}^{l \times m} \rightarrow \mathbb{R}_{A\md{H}B_1}^{k \times m}$, where $A\md{H}$, $B\md{H}$ are the input domains, $A\md{H}B_1$ is the output domain, and $l$ and $m$ are the dimensions of the input, $k$ and $m$ dimensions of output samples. 

The discriminator network is denoted as $\mathbf{D}_{A\md{H}}: \mathbb{R}_{A\md{H}B_1}^{k \times m} \rightarrow [0,1]$, and it discriminates samples in domain $A\md{H}B_1$. 
Each generator takes an observation of the size $l \times m$, and passes it to the encoder-decoder pair.
Note that $\mathbf{G}_{B\md{H}A_1}$, $\mathbf{G}_{A\md{H}B_2}$, $\mathbf{G}_{B\md{H}A_2}$, $\mathbf{D}_{B\md{H}}$, $\mathbf{D}_{A\md{H}B}$ and $\mathbf{D}_{B\md{H}A}$ are similarly defined.
The encoder and the decoder are composed of fully-connected layers. 
The number of layers ranges from 5 to 10 depending on a domain. 
The discriminator has an additional layer, a sigmoid function to output a predicted label.

\section{Experiments}
\label{sec:exp}

\subsection{Task Description: Exploring Novel Hydrides}

Hydrides, compounds which associate hydrogen atoms with other chemical elements, are actively used in storage battery technologies such as nickel-metal hydride battery. 
A number of hydrides have been explored as a means of hydrogen storage for fuel-cell powered electric cars.  

Crystallographic structures can be represented using the POSCAR files which are input files for the DFT calculations under the VASP code~\citep{Kresse99}. 
These are coordinate files, they contain the lattice geometry and the atomic positions, as well as the number (or the composition) and the nature of atoms in the crystal unit cell.

\begin{figure}
\centering
\includegraphics[scale=1.45]{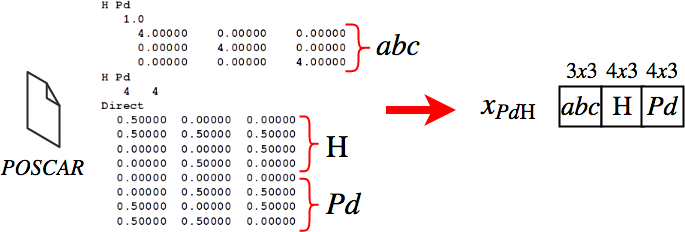}
\caption{An example of a POSCAR file describing the composition of Palladium and Hydrogen, and the data representation in the CrystalGAN.}
\label{fig:poscar}
\end{figure}

\begin{figure}
\centering
\includegraphics[scale=0.1]{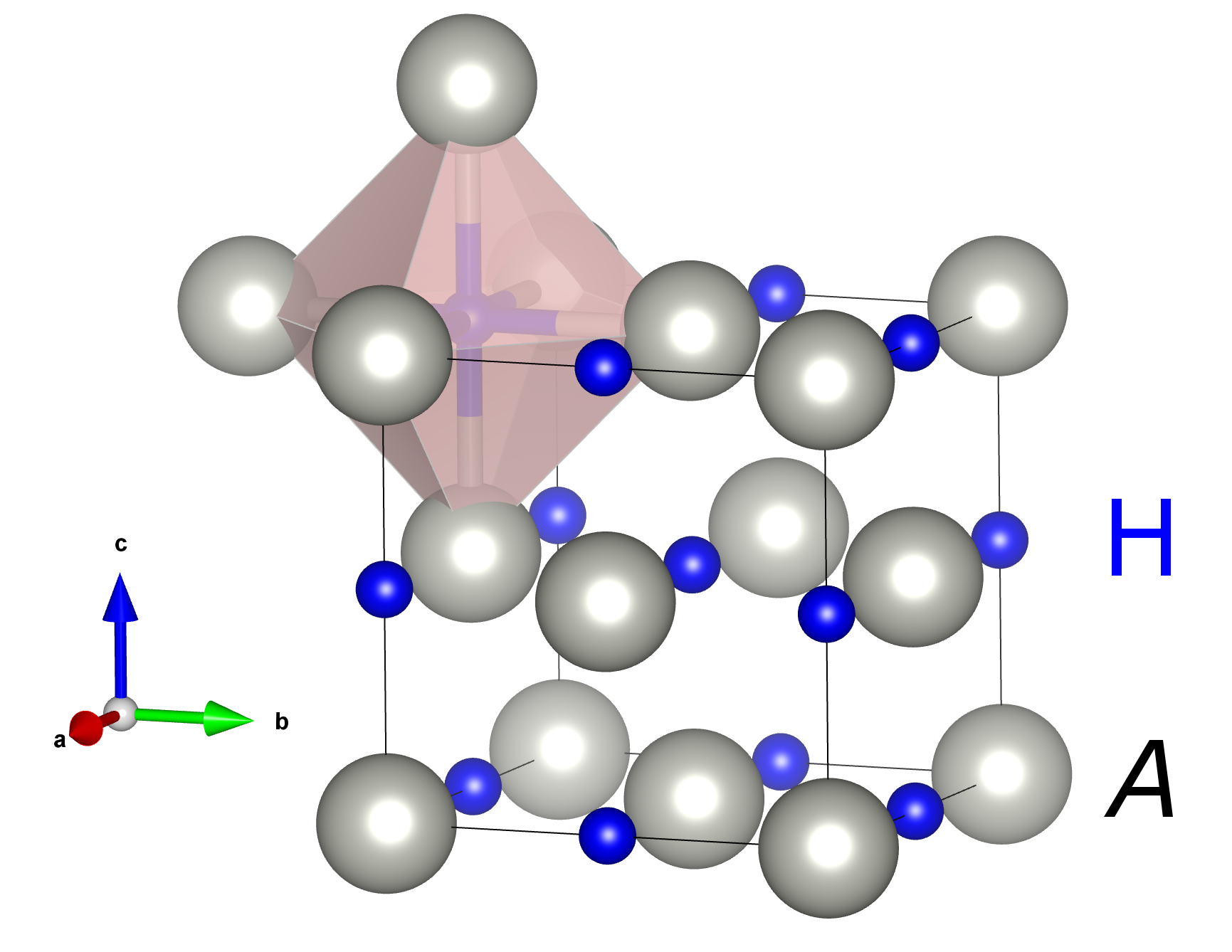}
\caption{A visualization of a stable structure.}
\label{fig:3dAtoms}
\end{figure}

We use a dataset constructed from~\citep{Bou17,Vil17} by experts in materials science. 
Our training data set contains the POSCAR files, and the proposed CrystalGAN generates also POSCAR files. 
Such a file contains three matrices: the first one is $abc$ matrix, corresponding to the three lattice vectors defining the unit cell of the system, the second matrix contains atomic positions of H atom, and the third matrix contains coordinates of metallic atom~$A$ (or~$B$). 
The information from the files is fed into 4-dimensional tensors. 
An example of a POSCAR file, and its corresponding representation for the GANs is shown on Figure~\ref{fig:poscar}. On Figure~\ref{fig:3dAtoms} we show the corresponding structure in 3D. 
Note that we increase the data complexity by the feature transfer procedure by adding placeholders.

\begin{figure}
\centering
\includegraphics[scale=0.345]{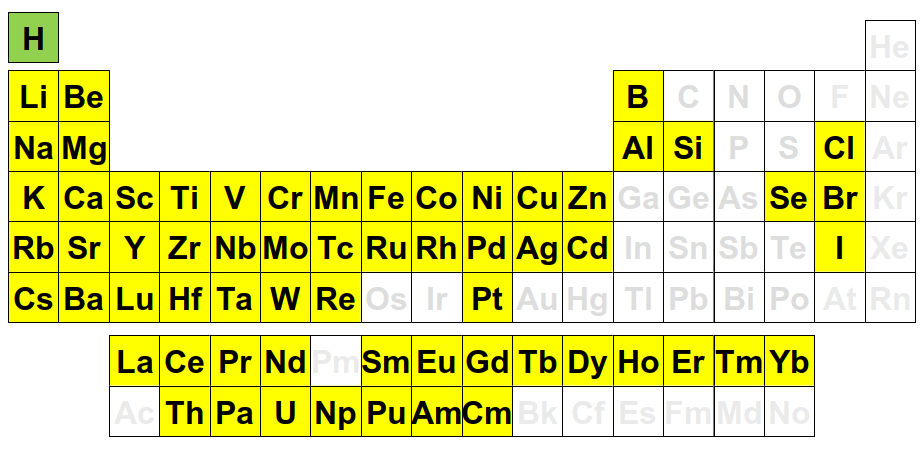}
\caption{The elements included in our data set are highlighted.}
\label{fig:periodictable2}
\end{figure}

Our training dataset includes 1,416 POSCAR files of binary hydrides divided into 63~classes where each class is represented as a 4-dimensional tensor.
Each class of binary $M$H hydride contains two elements: the hydrogen H and another element~$M$ from the periodic table.
This later is selected from the 63 highlighted $M$ elements (in yellow) in the Figure~\ref{fig:periodictable2}.
In our experiments, after discussions with materials science researchers, we focused on exploration of ternary compositions "Palladium - Hydrogen - Nickel" from the binary systems observations of "Palladium - Hydrogen" and "Nickel - Hydrogen". 
So, $A\md{H}$ = PdH, and $B\md{H}$ = NiH. We also considered another task to generate ternary compounds "Magnesium - Hydrogen - Titanium".

\begin{table*}
\centering
\begin{tabular}{|c|c|c|c|c|}
\hline
Composition & GAN & DiscoGAN & CrystalGAN  & CrystalGAN \\ 
& (standard) &  & without constraints  & with geometric constraints \\
\hline
Pd - Ni - H & 0 & 0 & 4 & 9 \\
\hline
Mg - Ti - H & 0 & 0 & 2 & 8\\
\hline
\end{tabular}
\caption{Number of ternary compositions of good quality generated by the tested methods.}
\label{tab:results}
\end{table*}

From each system (domain), we have selected 35~crystal structures (stable and metastable) which include experimentally observed prototypes. 
Here is a brief data description for this task:
\begin{center}
\begin{tabular}{|c|c|}
\hline
Input dataset & Dimension \\
\hline
PdH & $[35, 4, 18, 3]$\\
\hline
NiH & $[35, 4, 18, 3]$\\
\hline
\end{tabular}
\end{center}
where 18 and 3 are the maximal numbers of lines and columns in each matrix respectively.

In the CrystalGAN, we need to compute all the distances of the nearest neighbors for each generated POSCAR file. 
The distances between hydrogen atoms H in a given crystallographic structure should respect some geometric rules, as well as the distances between the atoms $A-B$, $A-A$', and $B-B'$. We applied the geometric constraints on the distances between the neighbors (for each atom in a crystallographic structure) introduced in the previous section.
Note that the distances $A-$H and $B-$H are not penalized by the constraints.

\subsection{Implementation Details}

In order to compute the distances between all nearest neighbors in the generated data, we used the pythonic library Pymatgen~\citep{Ong12} specifically developed for material analysis. 

For all experiments in this paper, the distances are fixed by our colleagues in crystallographic and materials science to $d_1 = 1.8$\,\AA\ (angstrom, $10^{-10}$\,meter) and $d_2 = 3$\,\AA. 
We set all the hyper-parameters by cross validation, however, we found that a reasonable performance is reached when all $\lambda_i$ have similar values, and are quite close to 1.
We use the standard AdamOptimizer with learning rate $\alpha = 0.0001$, and $\beta_1 = 0.5$. 
The number of epochs is set to 1000 (we verified that the functions converge). 
The mini-batch size equals~35.

Each block of the CrystalGAN architecture (the generators and the discriminators) is a multi-layer neural network with 5 hidden layers. 
Each layer contains 100 units. 
We use the rectified linear unit (ReLU) as an activation function of the neural network. 
All these parameters were fixed by cross-validation (for both chosen domains "Palladium - Hydrogen" and "Nickel - Hydrogen").

Our code is implemented in Python (TensorFlow).
We run the experiments using GPU with graphics card NVIDIA Quadro M5000.

\subsection{Results}

In our numerical experiments, we compare the proposed CrystalGAN with a classical GAN, the DiscoGAN~\cite{Kim17}, and the CrystalGAN but without the geometric constraints.
All these GANs generate POSCAR files, and we \textit{evaluate the performance of the models by the number of generated ternary structures which satisfy the geometric crystallographic environment}.
Table~\ref{tab:results} shows the number of successes for the considered methods.
The classical GAN which takes Gaussian noise as an input, does not generate acceptable chemical structures.
The DiscoGAN approach performs quite well if we use it to generate novel pseudo-binary structures, however, it is not adapted to synthesize ternary compositions.
We observed that the CrystalGAN (with the geometric constraints) outperforms all tested methods. 

\begin{figure*}
\centering
\includegraphics[scale=0.6]{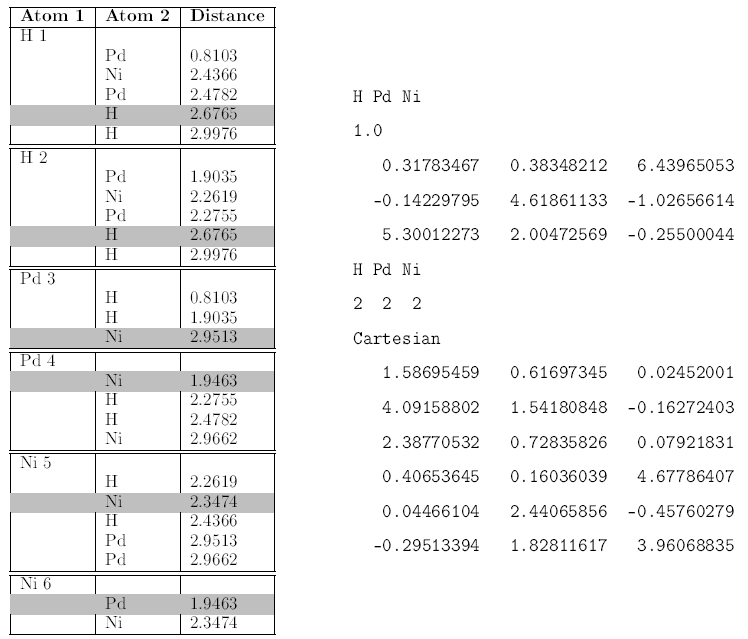}
\caption{The list of the nearest neighbours (on the left); the corresponding generated POSCAR file (on the right).}
\label{fig:stableStrPoscar}
\end{figure*}

Figure~\ref{fig:stableStrPoscar} illustrates characteristics of a newly generated ternary (H-Pd-Ni) stable structure: on the left we show the distances between the nearest neighbours in the crystallographic structure, and on the right we visualise the generated POSCAR file. We would like to underline that the generated structure respects the geometric constraints.



\section{Discussion}

Here we provide some important remarks on the task considered in this contribution. 
Discovery of stable chemical structures in general, and of new materials for hydrogen storage in particular, is a challenging task. 

From multiple discussions with experts in materials science and chemistry, first, we know that the number of novel stable compounds can not be very high, and it is already considered as a success if we synthesize several stable structures which satisfy the constraints. 
Hence, we can not really reason in terms of accuracy or error rate which are widely used metrics in machine learning and data mining.

Second, evaluation of a stable structure is not straightforward.
Given a new composition, only the result of density functional theory (DFT) calculations can provide a conclusion whether this composition is stable enough, and whether it can be used in practice. 
However, the DFT calculations are computationally too expensive, and it is out of question to run them on all data we generated using the CrystalGAN.
In our work, to avoid the DFT computations, we imply the geometric constraints proposed by the human experience to control the properties of the generated compounds, such as the Switendick criterion~\citep{Switendick}.
It is planned to run the DFT calculations on some pre-selected generated ternary compositions to take a final decision on practical utility of the chemical compounds. 

\begin{figure}
\centering
\includegraphics[scale=0.5]{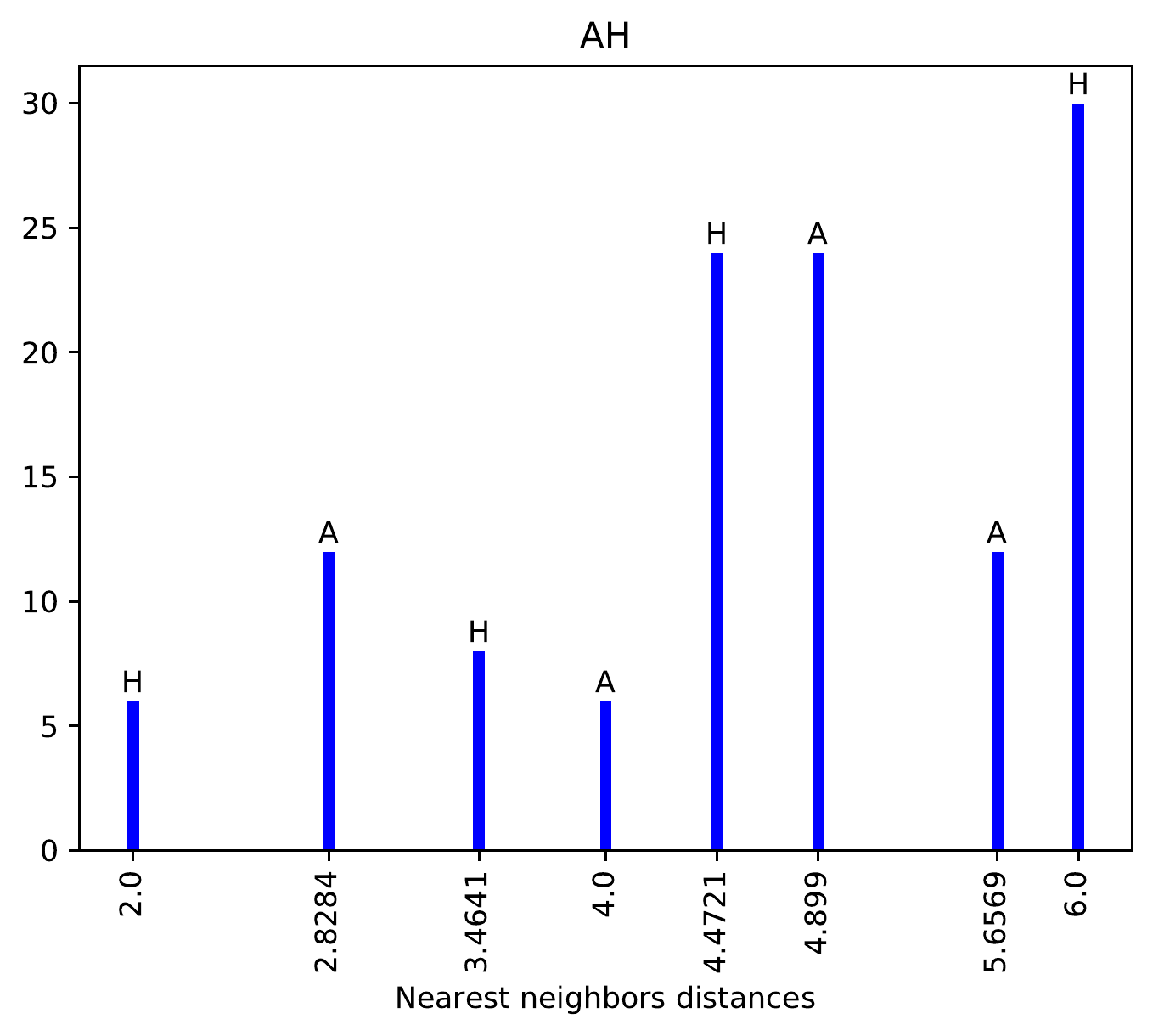}
\caption{Number of nearest neighbors at a given distance for each atom in a structure.}
\label{fig:nearest}
\end{figure}

The evaluation of generated crystallographic structures can also be done by laboratory experiments, exploring geometric properties of the compositions based on the distances between atoms. For example,~Figure~\ref{fig:3dAtoms} illustrates a stable structure in cubic NaCl prototype. 
Another representation of a synthesized data is a histogram of the number of nearest neighbors at a given distance which forms a pair distribution function (PDF). 
Figure~\ref{fig:nearest} shows a PDF~profile for a stable structure where the minimal distance between atoms is $d_\md{min}(A,\md{H})=2$\,\AA\ (angstrom) for 6~first nearest neighbours (cubic cell parameter is 4\AA\ in this example). 

\section{Conclusions}

Our goal was to develop a principled approach to generate new ternary stable crystallographic structures from observed binary, i.e. containing two chemical elements only. 
We propose a learning method called CrystalGAN to discover cross-domain relations in real data, and to generate novel structures. 
The proposed approach can efficiently integrate, in form of constraints, prior knowledge provided by human experts. 

CrystalGAN is the first GAN developed to generate scientific data in the field of materials science. 
To our knowledge, it is also the first approach which generates data of a higher-order complexity, i.e., ternary structures where the domains are well-separated from observed binary compounds. 
The CrystalGAN was, in particular, successfully tested to tackle the challenge to discover new materials for hydrogen storage.

Currently, we investigate different GANs architectures, also including elements of reinforcement learning, to produce data even of a higher complexity, e.g., compounds containing four or five chemical elements.
Note that although the CrystalGAN was developed and tested for applications in materials science, it is a general method where the constraints can be easily adapted to any scientific problem. 


%

\section*{Acknowledgements}
This work was supported by the French National Research Agency (ANR JCJC \textit{DiagnoLearn}).

\bibliography{GANs}
\bibliographystyle{aaai}

\end{document}